\documentclass[10pt,twocolumn,letterpaper]{article}

\usepackage[accsupp]{axessibility}  % Improves PDF readability for those with disabilities.
\usepackage{cvpr}      % To produce the REVIEW version
\usepackage{graphicx}
\usepackage{multirow}

\definecolor{cvprblue}{rgb}{0.21,0.49,0.74}
\usepackage[pagebackref,breaklinks,colorlinks,allcolors=cvprblue]{hyperref}

\def\paperID{5918} % *** Enter the Paper ID here
\def\confName{CVPR}
\def\confYear{2025}

\title{Align-KD: Distilling Cross-Modal Alignment Knowledge for \\ Mobile Vision-Language Large Model Enhancement}

\author{Qianhan Feng$\rm ^{1,2}\footnotemark[1]$ , Wenshuo Li$\rm ^{2}$, Tong Lin$\rm ^{1}\footnotemark[2]$ , Xinghao Chen$\rm ^{2}\footnotemark[2]$\\
$\rm ^{1}$State Key Laboratory of General Artificial Intelligence,\\ School of Intelligence Science and Technology, Peking University, China\\
$\rm ^{2}$Huawei Noah's Ark Lab, China\\
{\tt\small fengqianhan@stu.pku.edu.cn, lintong@pku.edu.cn, \{liwenshuo,xinghao.chen\}@huawei.com}}

\begin{document}
\maketitle
\renewcommand{\thefootnote}{\fnsymbol{footnote}}
% \footnotetext[1]{Work done during the internship at Noah's Ark Lab.}
% \footnotetext[1]{Corresponding authors.}
\begin{abstract}
Vision-Language Models (VLMs) bring powerful understanding and reasoning capabilities to multimodal tasks. Meanwhile, the great need for capable aritificial intelligence on mobile devices also arises, such as the AI assistant software. Some efforts try to migrate VLMs to edge devices to expand their application scope. Simplifying the model structure is a common method, but as the model shrinks, the trade-off between performance and size becomes more and more difficult. Knowledge distillation (KD) can help models improve comprehensive capabilities without increasing size or data volume. However, most of the existing large model distillation techniques only consider applications on single-modal LLMs, or only use teachers to create new data environments for students. None of these methods takes into account the distillation of the most important cross-modal alignment knowledge in VLMs. We propose a method called Align-KD to guide the student model to learn the cross-modal matching that occurs at the shallow layer. The teacher also helps student learn the projection of vision token into text embedding space based on the focus of text. Under the guidance of Align-KD, the 1.7B MobileVLM V2 model can learn rich knowledge from the 7B teacher model with light design of training loss, and achieve an average score improvement of 2.0 across 6 benchmarks under two training subsets respectively.

\end{abstract}    
\section{Introduction}
\label{sec:intro}

Vision Language Model (VLM) is an important Multimodal technology, which build a bridge between vision and text data, and facilitate many real world tasks and applications \cite{lin2020interbert,DBLP:conf/cvpr/FangW0LGWY022}. Based on the success of Large Language Models (LLMs) \cite{touvron2023llamaopenefficientfoundation,DBLP:journals/corr/abs-2307-09288,zhang2024tinyllama}, efforts have been done to integrate vision modal features with LLMs to extend models' capability and their application potential and build up new Vision-Language Models (VLMs) \cite{zhu2023minigpt,DBLP:conf/nips/LiuLWL23a,dong2024internlm,DBLP:journals/corr/abs-2401-15947,DBLP:journals/corr/abs-2301-13823}. However, new issues rise up: as the input features become more complex, the structures of VLMs also become deeper and heavier, since they have to digest information from different modalities and face even more various scenes \cite{bai2023qwen,chen2023sharegpt4v}. The growing size and complexity of VLMs makes them difficult to be accessed outside the server or high-speed Internet, which limit the development of these cutting-edge artificial intelligence under different scenarios, especially their deployment in off-line devices like mobile phones and robots, or some confidential application devices. 

\begin{figure}[t]
    \centering
    \includegraphics[width=\columnwidth]{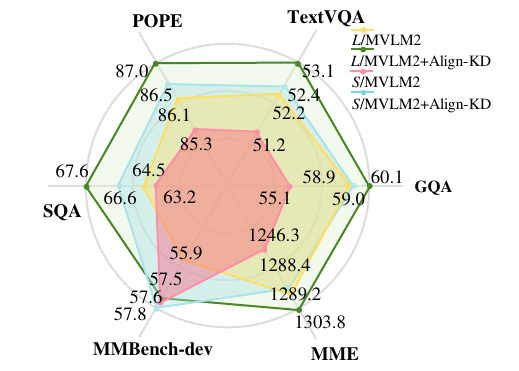}
    \caption{Radar plot of MobileVLM V2 1.7B model' performances with Align-KD policy under different settings. \textit{L} and \textit{S} refer to different \textit{Long} and \textit{Short} subdatasets for training, and MVLM2 refers to MobileVLM V2 model.}
    \label{fig:radar}
\end{figure}

Growing attentions have been focused on compressing VLMs while maintaining their remarkable capability as better as possible. MobileVLM family models \cite{DBLP:journals/corr/abs-2312-16886,DBLP:journals/corr/abs-2402-03766} are the first works to scale down VLMs to be able to run on mobile devices. Both MobileVLM V1 and MobileVLM V2 model contain special designed lightweight downsample vision projector to embed tokens into text embedding dimensions. With novel training strategy, MobileVLM V2 1.7B model outperforms a bunch of VLMs at 7B scale. Even so, continue to scale down the model would encounter more tough situations, where more severe performance drop would occur. In this case, new strategies are in need to aid the development of mobile VLMs.

In contrast to cutting down the scale of mobile VLMs with the risk of significant performance drop, we are motivated to boost the model without enlarging the scale of data amount. Knowledge distillation (KD) method is widely used to increase the capability of neural networks, aims to instruct smaller student model with a larger and stronger teacher model, thus learning teacher's behaviors or hidden representations \cite{DBLP:journals/corr/HintonVD15}. Previous knowledge distillation method for large model are mainly designed for single NLP modality \cite{DBLP:conf/iclr/Gu0WH24,DBLP:conf/icml/LiangZZHCZ23,DBLP:conf/acl/HsiehLYNFRKLP23}. In the field of vision-language multimodal models, most works are done before the large model era, focusing on aligning the vision proposals at the front side \cite{DBLP:conf/iccv/FangW0WY021}, which are no longer used in VLM technologies. Other works apply MSE loss on every transformer layers between models, but is not suitable for VLMs with significantly more layers with even different structures. 

Among these works, we notice that the alignment of vision and text inputs, the most important aspect of VLMs, is not considered in distillation. Poorly aligned cross-modal features could lead to difficulties in comprehending or reasoning. In this paper, we propose a lightweight knowledge distillation method, namely Align-KD, to let 1.7B student model learn the alignment knowledge from much stronger teachers. Firstly, we conduct several experiments on well-trained VLMs, and find that the first and last Transformer layer brings the largest shift on the features, similar as the trends in LLMs \cite{sun2024transformer}. This contributes helps us to develop our belief that the alignment of modalities mainly happens at the shallow layers, where the input embeddings are projected to high dimension space for comprehending and reasoning. Then, given that the natural cross-modal querying mechanism of Attention block, we let student mimic teacher's text-query-vision attention distribution at the first layer. What's more, considering that the importance of vision tokens in the queue are different according to different text prompts, we inject teacher's informative vision embeddings unbalancedly into student's vision projector's output. Finally, we follow latest LLM research \cite{DBLP:conf/iclr/Gu0WH24} to calculate reversed Kullback-Leibler divergence (R-KLD) between outputs to aid more general mean-seeking learning.  

We apply Align-KD policy to distill MobileVLM V2 1.7B, the state-of-the-art open-source VLM for mobile devices, from MobileVLM V2 7B teacher. We formulate two different subdatasets with increasing limitation of prompt maximum length, thus testing the effectiveness of Align-KD and VLMs under resource-limited scenarios. Following the multi-step training strategy of MobileVLM V2, Align-KD helps MobileVLM V2 1.7B model obtains universal promotion across 6 different benchmarks. The results show that Align-KD has great potential to help mobile VLMs to get enhancement with limited computation resource.  

The main contributions of our work are the followings:
\begin{itemize}
    \item We propose a knowledge distillation method Align-KD for mobile VLMs, which is the first work to distill the key cross-modal alignment knowledge.
    \item Align-KD helps cutting-edge MobileVLM V2 1.7B model obtains stable enhancement across different settings and benchmarks, largely facilitates the application of VLMs on edge devices.
    \item Align-KD doesn't rely on specific design of VLMs, and only requires light training designs, which gives it great potential to expand to various resource limited scenes.
\end{itemize}

\section{Related Works}
\label{sec:relatedwork}
\subsection{Large Models and Their Boosting}

\begin{figure*}[t]
    \centering
    \includegraphics[width=\textwidth]{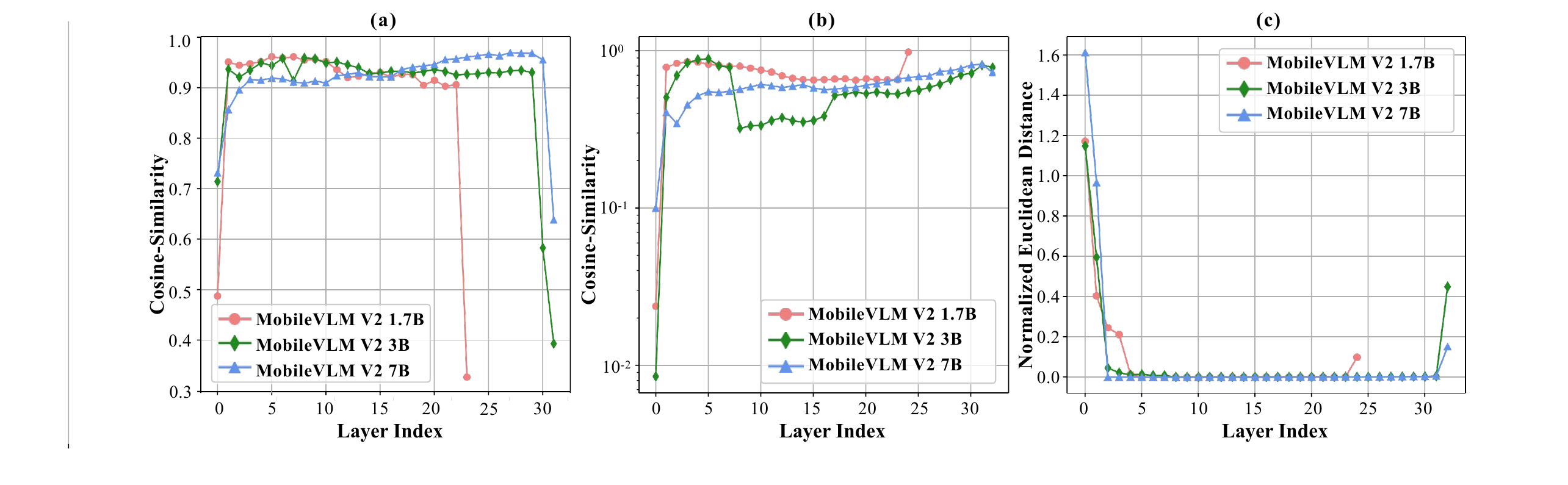}
    \caption{Exploration of feature changing trend at different layers of MobileVLM V2 model families. (a) Cosine similarity of features from every two adjacent layers. (b) Cosine similarity of features from original vision and text embedding positions within each same layer. The data is presented in an order of magnitude to highlight the trend of change. (c) Normalized Euclidean distance of features from original vision and text embedding positions within each same layer. The experiments are conducted on ShareGPT4V-PT dataset, and all the calculations are averaged after conducted on each tokens.}
    \label{fig:similarity}
\end{figure*}

In recent years, Large Language Models (LLMs) like GPT-3 \cite{brown2020language}, OPT \cite{DBLP:journals/corr/abs-2205-01068} and LLaMA \cite{touvron2023llamaopenefficientfoundation} significantly break through the borderline of deep learning and its applications. ChatGPT \cite{chatgpt} set off a new wave and inspired follow-up work such as Vicuna \cite{chiang2023vicuna}. Besides, some works try to introduce multimodal knowledge into the large model \cite{bai2023qwen,chen2023sharegpt4v,zhu2023minigpt,DBLP:journals/corr/abs-2410-13848,DBLP:journals/corr/abs-2501-17811,DBLP:journals/corr/abs-2411-07975,DBLP:journals/corr/abs-2412-10302}. LLaVA \cite{DBLP:conf/nips/LiuLWL23a} feed visual tokens into LLM and build up a comprehensive reasoning between visual and text contents, and many other works \cite{dong2024internlm,DBLP:journals/corr/abs-2401-15947,DBLP:journals/corr/abs-2301-13823,DBLP:journals/corr/abs-2312-06109} also approach to balance between vision and text understanding. However, the growing size of LLMs leads to a high demand on computing resources, which limits their applications. TinyLLaMA \cite{zhang2024tinyllama} and MobileLLaMA \cite{chu2023mobilevlm} scale down the architectures and maintain relatively good performance. Meanwhile in vision-language model field, MobileVLM family \cite{DBLP:journals/corr/abs-2312-16886,DBLP:journals/corr/abs-2402-03766} is the first open source work to facilitate the Vision-Language Model on mobile devices. Except for the development of training strategy and special architectures for large model, model compression techniques including quantization and pruning \cite{DBLP:conf/icml/FrantarA23,DBLP:journals/corr/abs-2210-17323} also thrive and provide solutions to relief burden of the computation resources. Equipped with these methods, LLMs are able to inference faster and lighter with little drop in accuracy.

\subsection{Large Language Model Distillation}

While former techniques are trying to do the subtraction, knowledge distillation (KD) \cite{DBLP:journals/corr/HintonVD15} techniques are trying to do adding. In KD, weaker student model tries to learn from a stronger teacher model from different aspects, like the output or hidden representations. MiniLLM \cite{DBLP:conf/iclr/Gu0WH24} studies the Kullback-Leibler divergence (KLD) loss on the output distillation, and suggests that reverse KLD teaches LLM student better on mean-seeking than forward KLD. Other researchers pay attention to the hidden features, training a task-aware filter to distill knowledge from teacher to student at different middle layers \cite{DBLP:conf/icml/LiangZZHCZ23}. Teachers can also be used to create a more suitable data environment for student model, especially in LLM background where data are diverse and sometimes polluted. Hsieh et al. \cite{DBLP:conf/acl/HsiehLYNFRKLP23} device a step-by-step distillation strategy to use teacher model's inference ability to provide training data for student, thus injecting both label noise and text inference into the data. Meta delves deeply into black-box systems \cite{yu2024distilling}, distilling high-quality outputs generated by System 2 techniques, such as Chain-of-Thought, Rephrase and Respond, etc., back into the standard large language model generation.

\subsection{Distillation for Vision-Language Model}

Most distillation methods for Vision-Language Model (VLM) are designed before Large Model Era. Considering that traditional VLMs rely on vision proposals, Fang et al. \cite{DBLP:conf/iccv/FangW0WY021} propose to align the input proposals between teacher and student, and enable following transformer blocks to align their attention distributions. To compress VLM, Wang et al. \cite{DBLP:conf/acl/WangZ0Z23} combine pruning with distillation, conducting easy output logits imitation and distillation on attention and hidden states. Although these works provide a thinking of VLM distillation, they are restrained within the field outside Vision-Language Model (VLM), which usually consists of more transformer layers and more complex alignment between vision and language modalities. In VLM-KD \cite{DBLP:journals/corr/abs-2408-16930}, researchers use VLM like LLaVA-NeXT \cite{liu2024llavanext} to generate text prompts and using contrastive learning to promote long-tail recognition ability of vision models. LLaVA-MoD \cite{shu2024llava} minimizes the Kullback-Leibler divergence between output distributions and utilizes Direct Preference Optimization (DPO) to enhance the ability of the s-MLLM to discern high-quality from low-quality samples. But LLaVA-MoD relies on integrating the Sparse Mixture of Experts (MoE) architecture into language model, and also neglects the gap of alignment knowledge in distillation, which leaves a huge space for further explorations.  
\section{Align-KD}

In Vision-Language Models (VLMs), vision and text embeddings comprise the input of the large model. However, it is obvious that the embedding mechanisms are different for the two modalities, which means the embeddings have to go through cross-modal alignment in the feature space. The cross-modal alignment ability is crucial for VLMs, but previous works mainly focus on single modal LLMs distillation and neglect the importance of teaching student about the alignment knowledge. Here we first explore the cross-modal alignment in VLMs, and then propose our Align-KD method step by step based on MobileVLM family. 

\begin{figure*}[t]
    \centering
    \includegraphics[width=\textwidth]{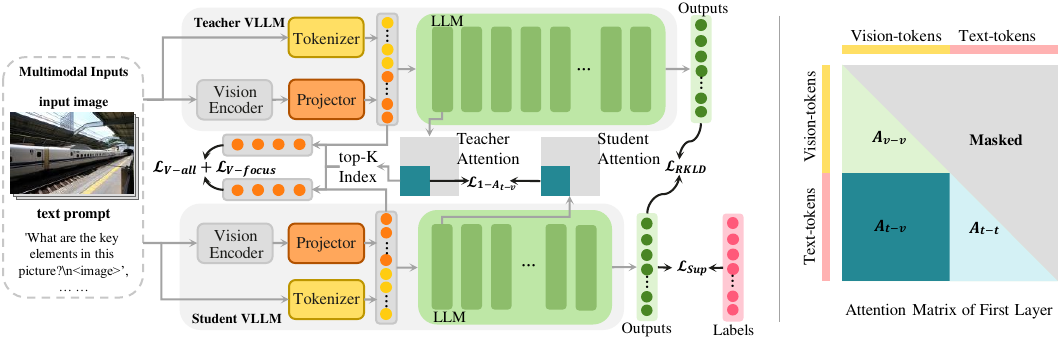}
    \caption{\textbf{Left:} The overall framework of Align-KD. Align-KD utilizes the text-query-vision attention of teacher model's first layer to extract the knowledge of cross-modal alignment, then injects this knowledge into the cross-modal attention matrix of student's first layer. Besides, the projected vision tokens are dynamically enhanced according to text's focusing, also based on teacher's first layer cross-modal attention. \textbf{Right:} The schematic diagram of vision-language models' (VLMs) first layer attention matrix. 
    $A_{v-v}$, $A_{t-v}$ and $A_{t-t}$ attention refer to vision-query-vision, text-query-vision and text-query-text attention.}
    \label{fig:main}
\end{figure*}

\subsection{Where Does the Alignment Happen?}
\label{sec:discussion}

Most VLMs like MobileVLM \cite{DBLP:journals/corr/abs-2312-16886,DBLP:journals/corr/abs-2402-03766} design special vision projectors to project the vision embeddings, but this operation mainly align the dimension of embedded tokens. The alignment of vision and text embeddings into the same high dimension space is almost a black box system. 

Sun et al. \cite{sun2024transformer} try to figure out the internal working mechanism of Transformer layers in LLMs. The researchers perform both skip and switch operations on every Transformer layer in LLaMA 2 7B, 13B and 70B \cite{DBLP:journals/corr/abs-2307-09288} models, and find out that the change in the first and last layer brings the largest drop in performance, while middle layers only give slight fluctuation. What's more, the parameter weight of middle layers shows high similarity while the first and last layer are quite different. They conclude that the first and last layer take up special responsibilities.

This inspire us to explore whether the first and last layer in LLM of VLMs have similar functions to project input embeddings into/out of the high dimension space where the features mix. If so, the alignment of modalities should also happen simultaneously. Given the well trained MobileVLM V2 models with different sizes, we conduct 3 simple experiments to explore the feature change trend in different Transformer layers. First, we calculate the cosine similarity of every two adjacent features, including the input features. As shown in Figure \ref{fig:similarity}(a), the similarities in middle layers are much higher than the first and last layers, which implies dramatic change in feature space in the two layers. Then, we locate the positions of vision tokens and text tokens in the input embeddings, and calculate the feature cosine similarity as well as the Euclidean distance of these two segments within every layers, and the results are shown in Figure \ref{fig:similarity}(b) and Figure \ref{fig:similarity}(c). The experimental results further confirm our conjecture that the head layer of LLM maps the input to the high-dimensional space for deep processing, and the last layer maps it to the output space. This means that the first layer is also responsible for mapping text and vision embeddings from different domains into the same high-dimensional space, or alignment. Note that the LLMs in MobileVLM V2 1.7B model and 3B model are from MobileLLaMA family, and the LLM in 7B model is Vicuna-7B \cite{DBLP:conf/nips/ZhengC00WZL0LXZ23}, so the results are not structure-related. 

Except for the first layer in LLM of VLMs, models like MobileVLM also design vision projectors to downsample the vision embeddings, thus reducing the calculation. The projectors learn to maintain and enhance important information, and align the vision embeddings with the input text embeddings from the aspect of dimension. It also receives guidance from the backward gradient from downstream LLM to learn basic cross-modal knowledge.

Based of the discussion, we propose to distill the knowledge of cross-modal alignment from two aspects of VLMs: the first layer of LLM in VLMs, and the output embeddings of vision projectors as shown in the left of Figure \ref{fig:main}. 

\subsection{First Layer Text-Query-Vision Attention Only}

The first layer of LLM in VLMs is the important place where the cross-modal alignment happens. Almost all of recent LLMs are built based upon the architecture of Transformer \cite{DBLP:conf/nips/VaswaniSPUJGKP17}, an efficient parallel attention structure. While some works have been done to improve the Transformer block \cite{linrec,shen2018reinforced,sukhbaatar2019adaptive}, the basic attention mechanism remains as: project the input features into query ${Q}$, key ${K}$ and value ${V}$, and then use them to generate ${Attention}$ values to help self-adaptive fusion among different feature tokens. 

${Attention}$ values imply tokens' unbalanced focusing degree on others and determine how the input features are going to be projected. This nature of attention mechanism makes it the perfect information carrier about the cross-modal alignment in VLMs. The attention matrix of VLMs are always in the similar mode as the right of Figure \ref{fig:main}, where the input embeddings are the concatenated vision and text tokens. Since VLMs mainly use decoder transformer layers, the attention matrix is a lower triangular matrix and half masked. The lower part of the matrix consists of three parts: vision-query-vision attention $A_{v-v}$, text-query-text attention $A_{t-t}$ and text-query-vision attention $A_{t-v}$.

Previous knowledge distillation methods usually teach student model to learn from the whole attention values of teacher's different layers. On the contrary, we propose to let student model mimic the first layer's text-query-vision attention ($A_{1,t-v}$) only. Given a knowledgable teacher model $T$, we pick out its attention matrix $A^{T}_{1}$ from its first layer, then split out the text-query-vision part to form $A_{1,t-v}^{T}$ and do the same to get student $S$'s first layer's text-query-vision attention $A_{1,t_v}^{S}$. Then, we apply an ${1\times1}$ convolution projector ${P_{attn}}$ to align the dimension of these two, followed by a simple mean square error (MSE) loss to get the first layer $A_{1,t-v}$ only KD loss:  
\begin{equation}
    \mathcal{L}_{A_{1,t-v}} = MSE(P_{attn}(A_{1,t-v}^{T}),A_{1,t-v}^{S}).
\end{equation}

This first layer $A_{1,t-v}$ only KD has several advantages. (1) First and foremost, thanks to the cross-attention mechanism, the text-query-vision attention matrix of the first layer naturally implies how the input tokens of two modalities perceive each other, as well as the alignment projection scheme that tokens are going to take to project into more aligned feature space. On the contrary, it is difficult to extract useful knowledge from $A_{1,t-t}$ when then text tokenizer of VLMs are usually fixed, while distilling $A_{1,v-v}$ only lead to more vision-only enhancement, which has already been reinforced by front vision projector. And both of them lack the key cross-modal knowledge. (2) Secondly, first layer $A_{1,t-v}$ only KD significantly reduces computing workload, making the whole method efficient. Some works distill attention matrixes at multiple layers, and some of them even device special downstream tasks to help the student learning, which can cause excessive computational pressure. Even within the first attention matrix, first layer $A_{1,t-v}$ only KD also saves up to 50\% calculation compared with full distillation. This lightweight design enable potential chances to conduct training with limited computation resources. (3) What is more, considering that the design of different VLMs could be different in both the model depth and the block details, first layer $A_{1,t-v}$ only KD exhibits greater flexibility and can be easily migrated to different models regardless of their particular structural design or the meaning of different depth transformer layers.

\begin{figure}[t]
    \centering
    \includegraphics[width=\columnwidth]{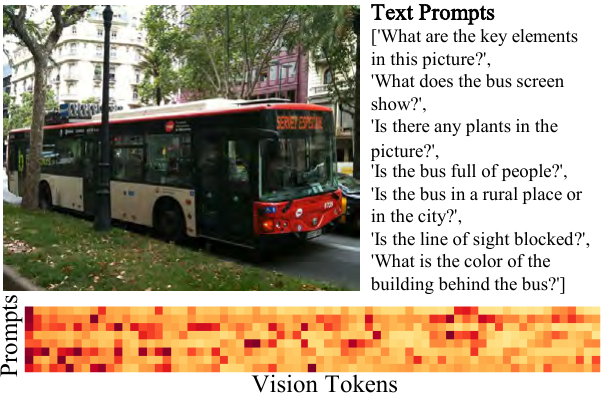}
    \caption{Different text prompts cause different attention on vision tokens of the picture. The vision tokens with high attention activation also distribute sparsely.}
    \label{fig:heatmap}
\end{figure}

\subsection{Vision Enhancement Based on Text's Focusing}

The vision projector that generates the input vision tokens also takes up the responsibility to do rudimentary cross-modal alignment. For example, the LDP in MobileVLM model downsample the embeddings while extracting both detail and semantic features, and project the embeddings into less tokens with same dimension as text embeddings. This rough alignment lacks the perception of text modal information, only receiving backward gradients from downstream Transformer layers' cross-attention to get in touch with another modality. 

To mitigate this shortage, we propose to leverage the cross-modal attention to help the projector get more exposure to cross-modal information. Although the vision tokens are already refined by the projector, the text's attention on different vision tokens is still rather sparse because of the directional indication of text prompts. The $A_{1,t-v}$ not only contains information about cross-modal aligning projection, but also reveals which vision tokens the text prompts pay most attention to. Different prompts may focus on different embeddings, but some of the tokens are significantly left behind. Instead of inhibiting the learning of temporarily unpopular vision tokens, we propose to enhance most popular tokens at current based on teacher model's $A_{1,t-v}$, thus preventing hurting the learning of others. Having the text-query-vision attention from the first layer of teacher model, we add-up the attention value ${A_{1,t-v}^{T}}$ along the text dimension to get the attention score $Score_{n}$ of vision token ${n}$,
\begin{equation}
    Score_{n} = \sum^{M}A^{T}_{1,t-v,(n,m)}, m \in M,
\end{equation}
$M$ is the number of text tokens. Then, the indexes of tokens whose attention score $Score_{n}$ if the top-$K$ are sorted out, namely $Idx_{K}$. Finally, we conduct knowledge injection from teacher to student on vision tokens listed in $Idx_{K}$: 
\begin{equation}
    \mathcal{L}_{V-focus}=MSE(P_{V}(Emb_{Idx_{K}}^{T}),Emb_{Idx_{K}}^{S}),
\end{equation}
where $Emb_{Idx_{K}}^{T}$ and $Emb_{Idx_{K}}^{S}$ are teacher's and student's vision token embeddings within the range of $Idx_{K}$, and $P_{V}$ is an ${1\times1}$ convolution projector.   

Except for the knowledge injection on current popular tokens, the rest should not be overlooked. Low-ranked attention from the current text prompt does not mean the stable low popularity under other scenes. The focus on vision tokens may dramatically shift, e.g., when from \textit{`What's the name of book's author on the cover?'} to \textit{`What is the color of the desk under the book?'}. What is more, the teacher vision token naturally contains stronger knowledge because of the instruction of stronger LLM and higher dimension in most cases. In this case, we add the general knowledge distillation on all vision tokens from teacher to student, i.e.,
\begin{equation}
    \mathcal{L}_{V-all}=MSE(P_{V}(Emb_{T}),Emb_{S}),
\end{equation}
which prevents harmful suppression of other tokens. And we combine two losses using a weight $\lambda$ to form the vision token enhancement loss based on text's focusing as
\begin{equation}
    \mathcal{L}_{V}=\mathcal{L}_{V-all}+\lambda \mathcal{L}_{V-focus}.
\end{equation}

\subsection{Overall Knowledge Distillation Strategy}

Except for alignment distillation in the front of VLM, we follow MiniLLM to add reverse Kullback-Leibler divergence (RKLD) loss, which uses the predicted distribution of student model as target, between the outputs of student and teacher. Since the text prompts and input images are varied, RKLD loss is more suitable than forward Kullback-Leibler divergence (FKLD) for VLM to learn about the mean-seeking instead of mode-seeking, preventing from overfitting on specific scene. Extract the output prediction distribution $p_{T}$, $p_{S}$ from teacher and student, the reverse Kullback-Leibler divergence loss can be formulated as
\begin{equation}
    \mathcal{L}_{RKLD}=p_{S}\log{\frac{p_{S}}{p_{T}}}.
\end{equation}

The alignment loss and RKLD loss work together with the original supervised loss $\mathcal{L}_{Sup}$, and our overall Align-KD loss on student model can be formulated as: 
\begin{equation}
    \mathcal{L} = \mathcal{L}_{Sup}+\mathcal{L}_{A_{1,t-v}}+\mathcal{L}_{V}+\mathcal{L}_{RKLD}.
\end{equation}
\section{Experiments}
\label{sec:experiments}

\subsection{Basic Experiment Settings}

MobileVLM family \cite{DBLP:journals/corr/abs-2312-16886,DBLP:journals/corr/abs-2402-03766} is the latest and cutting-edge Vision-Language Model for mobile devices. The MobileVLM V2 1.7B model shows remarkable performance with lightweight design. Considering the limitation of further compression of the model, we propose to apply Align-KD on MobileVLM V2 1.7B to help obtain better performance on edge devices. We choose the well trained MobileVLM V2 7B model as the teacher to conduct the knowledge distillation. Align-KD is used as an extra strategy working together with the common student training.

We use 8 NVIDIA V100 GPUs to conduct our distillation training. We follow MobileVLM V2's work to divide the training stage into pre-training and multi-task finetuning. Because of the limited memory space, we use gradient accumulation to achieve a global batch size of 256 for the pre-training stage and 128 for the multi-task fintuning stage. For both stage, we use the ZeRO2 strategy of DeepSpeed \cite{aminabadi2022deepspeed}, and run all experiments under half-precision floating-point. We follow MobileVLM V2 to freeze the vision encoder and tokenizer while training, only train rest of the whole network. The maximum learning rate for the projector and other components are set to $1e^{-3}$ and $2e^{-5}$ respectively during pretraining and $4e^{-5}$ during multi-task finetuning, using a cosine schedule. The convolutional projector for dimension alignment are randomly initialized and trained together. The weight $\lambda$ in vision token loss $\mathcal{L}_{V}$ is set to 0.1 to adapt to unstable changing of the tokens been focused. For top-$K$ selection in text-focus-based vision token knowledgable distillation, we select the top 16 tokens with highest attention score. We follow the original MobileVLM V2 work on the rest of settings. 

\begin{table*}[t]
    \centering
    \resizebox{0.95\textwidth}{!}{
    \begin{tabular}{ccc|ccccccc}
        \toprule[1pt]
        Method & LLM & \#Samples & MME$\rm ^{P}$ & GQA & VQA$\rm ^{T}$ & POPE & MMB$\rm ^{dev}$ & SQA$\rm ^{I}$ & Avg.\\
        \midrule[0.5pt]
        MiniGPT-4 & Vicuna-7B & 5.0M & 581.7 & 32.2 & - & - & 23.0 & - & - \\
        LLaVA-1.5\cite{DBLP:conf/cvpr/LiuLLL24} & Vicuna-7B & 1.2M & 1510.7 & 62.0 & 58.2 & 85.9 & 64.3 & 66.8 & 68.8 \\
        ShareGPT4V & Vicuna-7B & 1.9M & 1567.4 & 63.3 & 60.4 & 85.7 & 68.8 & 68.4 & 70.8 \\
        MoE-LLaVA-1.6B$\times$4 & StableLM-1.6B & 2.2M & 1300.8 & 60.4 & 47.8 & 84.3 & 59.4 & 62.6 & 63.3 \\
        MoE-LLaVA-2.7B$\times$4 & Phi-2.7B & 2.2M & 1396.4 & 61.1 & 50.2 & 85.0 & 65.5 & 68.7 & 66.7 \\
        MobileVLM 1.7B & MobileLLaMA 1.4B & 3.6M & 1196.2 & 56.1 & 41.5 & 84.5 & 53.2 & 57.3 & 58.7 \\
        MobileVLM 3B & MobileLLaMA 2.7B & 3.6M & 1288.9 & 59.0 & 47.5 & 84.9 & 59.6 & 61.2 & 62.8 \\
        MobileVLM V2 3B & MobileLLaMA 2.7B & 3.6M & 1440.5 & 61.1 & 57.5 & 84.7 & 63.2 & 66.7 & 68.1 \\
        MobileVLM V2 7B & Vicuna-7B & 3.6M & 1559.0 & 62.6 & 62.3 & 86.6 & 69.2 & 74.7 & 72.2 \\
        %LLaVA-MoD-2B & Qwen-1.5-1.8B & 5.0M & 1352.0 & 58.7 & 58.5 & - & 66.3 & 68.0 & - \\
        \midrule[1pt]
        \midrule[1pt]
        Subset / \#Samples & Student Model & Align-KD & MME$\rm ^{P}$ & GQA & VQA$\rm ^{T}$ & POPE & MMB$\rm ^{dev}$ & SQA$\rm ^{I}$ & Avg.\\
        \midrule[0.5pt]
        \multirow{2}*{\textit{Short} / 3.6M} & MobileVLM V2 1.7B & - & 1246.3 & 55.1 & 51.2 & 85.3 & 57.6 & 63.2 & 62.4 \\ 
        ~ & MobileVLM V2 1.7B & \checkmark & \textbf{1288.4} & \textbf{58.9} & \textbf{52.4} & \textbf{86.5} & \textbf{57.8} & \textbf{66.6} & \textbf{64.4} \\ 
        \midrule[0.5pt]
        \multirow{2}*{\textit{Long} / 3.6M} & MobileVLM V2 1.7B & - & 1289.2 & 59.0 & 52.2 & 86.1 & 55.9 & 64.5 & 63.7 \\ 
        ~ & MobileVLM V2 1.7B & \checkmark & \textbf{1303.8} & \textbf{60.1} & \textbf{53.1} & \textbf{87.0} & \textbf{57.5} & \textbf{67.7} & \textbf{65.1} \\ 
        \bottomrule[1pt]
    \end{tabular}
    }
    \caption{Test of Align-KD strategy's effectiveness on MobileVLM V2 1.7B model. \textit{Long} and \textit{Short} refer to two subsets with different maximum prompt lengths limitations. MME$\rm ^{P}$ refers to MME Perception, MMB$\rm ^{dev}$ refers to MMBench-dev, SQA$\rm ^{I}$ refers to SQA-IMG. The score of MME$\rm ^{P}$ is divided by 20 when calculating the average performance.}
    \label{main}
\end{table*}

\begin{table}[ht]
    \centering
    \resizebox{0.45\textwidth}{!}{
    \begin{tabular}{lll}
        \toprule[1pt]
        Datasets & \textit{Long} Samples & \textit{Short} Samples\\
        \midrule[0.5pt]
        \textit{\textbf{Pretraining}} & & \\
        ShareGPT4V-PT & 1.25M  & 1.24M\\
        \midrule[0.5pt]
        \textit{\textbf{Multi-task Finetuning}} & & \\
        COCO & 592K & 589k \\
        SBU & 837K* & 822k* \\
        Visual Dialog & 123K & 115K \\ 
        ShareGPT4V & 665K & 655K\\
        SQA & 13K & 5K \\
        IConQA & 107K & 107K\\
        TextVQA & 35K & 33K\\
        VSR & 13K & 13K\\
        VIGC & 37K & 35K\\
        \midrule[0.5pt]
        \textit{\textbf{Total}} & 3.67M & 3.61M\\
        \bottomrule[1pt]
    \end{tabular}
    }
    \caption{Details of datasets used in different stages. *SBU dataset is re-collected, some datas are removed from the original links.}
    \label{table:data}
\end{table}

\subsection{Data and Dataset Reforming}

Align-KD follows MobileVLM V2 to train on various datasets. During the pretraining stage, ShareGPT4V-PT \cite{chen2023sharegpt4v} is used to give the student a brief knowledge of vision and text. It is a caption dataset and comprises 1.2 million image-text pairs. In the multi-task training stage, more data from different tasks like conversation and VQA are provided: COCO\cite{DBLP:journals/corr/ChenFLVGDZ15}, SBU\cite{DBLP:conf/nips/OrdonezKB11}, Visual Dialog\cite{das2017visual}, ShareGPT4V\cite{chen2023sharegpt4v}, SQA\cite{DBLP:conf/nips/LuMX0CZTCK22}, IConQA\cite{DBLP:conf/nips/LuQCXZZYLZ21}, TextVQA\cite{DBLP:conf/cvpr/SinghNSJCBPR19}, VSR\cite{DBLP:journals/tacl/0001EC23}, VIGC\cite{DBLP:conf/aaai/WangWHPZZDLLWH24}. Note that SBU is a re-collected dataset and is updated from time to time, therefore some of the original data might have been removed. In this case, we washed the data list in the dataset, which is different from the original MobileVLM training. To evaluate the effectiveness of our Align-KD method, we test the performance  on different benchmarks, including GQA\cite{DBLP:conf/cvpr/HudsonM19}, SQA\cite{DBLP:conf/nips/LuMX0CZTCK22}, TextVQA\cite{DBLP:conf/cvpr/SinghNSJCBPR19}, MME\cite{DBLP:journals/corr/abs-2306-13394}, MMBench\cite{DBLP:conf/eccv/LiuDZLZZYWHLCL24} and POPE\cite{DBLP:conf/emnlp/LiDZWZW23}. 

While mobile VLMs are designed for mobile devices deployment with limited resources, the training of VLMs also faces challenges when the computational resources is not that adequate. The computation workload can be extremely high when the input text prompts are too long, which sometimes causes 'Out of Memory' error during training. We formulate two different subdatasets based on the original data listed above, each contains data with different maximum prompt lengths: \textit{Short} with maximum lengths of 512 embedded tokens, and \textit{Long} of 2048. Considering that the training data comprises many Visual Question Answering (VQA) tasks, we drop the overlong data instead of truncating them, making two subsets also vary in the data amount. This setting is different from the original MobileVLM work, but can help examine the effectiveness of Align-KD strategy under different resource-limited scenarios, and also helps extend the application scenarios of VLMs. The subset details are shown in Table \ref{table:data}.

\subsection{Effectiveness of Align-KD}

After formulating our \textit{Long} and \textit{Short} subdatasets, we use them to train MobileVLM V2 1.7B model with proposed Align-KD strategy. Note that we use fully-trained and open-sourced MobileVLM V2 7B model provided by MobileVLM work as our knowledge distillation teacher. The results across 6 different benchmarks and two subsets are shown in Table \ref{main}. Trained with \textit{Long} set, MobileVLM V2 1.7B model achieves an average score of 63.7. When trained with Align-KD policy, the student model witnesses a universal promotion across all benchmarks, achieving and average score of 65.1. To be more specific, Align-KD helps improve the MobileVLM V2 1.7B model to obtain an improvement of 3.2 on SQA benchmark, as well as 1.6 on MMBench. Under resource-limited scenarios, Align-KD gives MobileVLM V2 1.7B model an even better average improvement of 2 across all benchmarks, from 62.4 to 64.4. On GQA, Align-KD gives a promotion of 3.8. And it also brings notable promotion of 1.2 on POPE, which is a rather challenging hallucination testing benchmark. 

The comprehensive results are visualized in a radar plot in Figure \ref{fig:radar}. Our Align-KD brings stable and good bonus under different vision-language tasks. What's more, when the student suffers from the performance drop brought by the absence of long prompts, Align-KD successfully injects the knowledge into the model and makes student model after knowledge distillation achieve performances comparable with the model trained with long texts.

\subsection{Ablation Study}

We take a step forward to conduct ablation studies to testify the effectiveness of each method in Align-KD. We run all ablation experiments on \textit{Short} subdataset for fairness, and the results are shown in Table \ref{table:ablation-main}. The reverse Kullback-Leibler divergence (RKLD) loss helps student learn to mimic the outputs of stronger teacher model, but brings somehow biased improvement. On the contrary, applying first layer $A_{1,t-v}$ only loss provides notable and more balanced promotion to an average of 63.6. The combination of distillation on focused and all vision tokens further increase the performance by 0.8 on average. 

\begin{table}[t]
    \centering
    \setlength{\tabcolsep}{2pt}
    \resizebox{\columnwidth}{!}{
    \begin{tabular}{lccccccc}
        \toprule[1pt]
        Method & MME$\rm ^{P}$ & GQA & VQA$\rm ^{T}$ & POPE & MMB$\rm ^{dev}$ & SQA$\rm ^{I}$ & Avg.\\
        \midrule[0.5pt]
        MVLM2 1.7B & 1246.3 & 55.1 & 51.2 & 85.3 & 57.6 & 63.2 & 62.4 \\
        +$\mathcal{L}_{RKLD}$ & 1223.2 & 55.9 & 52.5 & 85.5 & 58.5 & 64.7 & 63.0 \\ 
        +$\mathcal{L}_{A_{1,t-v}}$ & 1263.6 & 57.8 & 53.1 & 85.7 & 57.2 & 64.6 & 63.6 \\
        +$\mathcal{L}_{V-all}$ & 1288.2 & 57.7 & 52.3 & 86.2 & 57.8 & 64.3 & 63.8 \\
        +$\mathcal{L}_{V-focus}$& 1288.4 & 58.9 & 52.4 & 86.5 & 57.8 & 66.6 & 64.4 \\
        \bottomrule[1pt]
    \end{tabular}
    }
    \caption{The ablation results of testing each component in Align-KD. MVLM2 refers to MobileVLM V2 model.}
    \label{table:ablation-main}
\end{table}

During cross-modal alignment learning, we apply the knowledge distillation only on the text-query-vision part of the first layer's attention. We further testify the rationality of this design by comparing with other methods, and the results are shown in Table \ref{table:ablation-attention}. Replacing first layer $A_{1,t-v}$ only KD strategy with distillation of vision-query-vision attention triggers extreme drop in performances. We suggest that this is because the self-attention of vision tokens is not helpful for initial cross-modal alignment, and could lead to very unstable fluctuation in the attention embedding. However, changing into text-query-text attention is remarkably better than vision-query-vision training, with a slight drop from 64.4 to 63.2. We believe this phenomenon can be attributed to the fixed text tokenizer while training, and this lead to indirect stable learning of projection of cross-modal attention. What's more, the learning on full attention performs better than learning only $A_{1,v-v}$ and $A_{1,t-t}$, which further improves our conjectures above: the cross-modal $A_{1,t-v}$ is the most important factor while learning $A_{1,t-t}$ helps mitigate the negative influence brought by $A_{1,v-v}$ distilling.

In Figure \ref{fig:similarity}, we infer that the first and last layer act as the most outstanding characters. Align-KD mainly focuses on the first layer, since it is exposed the most to the inputs that need to be aligned, while the last layer is mainly in charge of projecting into output space. However, we also testify how the student would behave with the knowledge of output attention. As it is shown in the last row of Table \ref{table:ablation-attention}, after adding the distillation of the last attention to Align-KD, student model witnesses a drop in performance. A possible explanation is that the deep feature is already well mixed and extracted, and there is some overlap with functionality of the RKLD on outputs. Besides, the distillation on extra attention would lead to significant growth in the calculation, which is a burden for resource-limited scenarios, and the first layer $A_{1,t-v}$ only KD in Align-KD demonstrates huge advantage in both effectiveness and efficiency.

\begin{table}[t]
    \centering
    \setlength{\tabcolsep}{2pt}
    \resizebox{\columnwidth}{!}{
    \begin{tabular}{lccccccc}
        \toprule[1pt]
        Method & MME$\rm ^{P}$ & GQA & VQA$\rm ^{T}$ & POPE & MMB$\rm ^{dev}$ & SQA$\rm ^{I}$ & Avg.\\
        \midrule[0.5pt]
        $A_{1,t-v}$ & 1288.4 & 58.9 & 52.4 & 86.5 & 57.8 & 66.6 & 64.4 \\
        $A_{1,v-v}$ & 1036.5 & 52.0 & 27.3 & 81.1 & 13.8 & 36.1 & 43.7 \\
        $A_{1,t-t}$ & 1228.2 & 61.0 & 49.7 & 86.2 & 56.4 & 64.3 & 63.2 \\
        $A_{1,all}$ & 1265.0 & 59.8 & 52.7 & 86.3 & 54.4 & 64.6 & 63.5\\
        $+A_{last,all}$ & 1281.1 & 58.7 & 53.8 & 86.1 & 57.7 & 64.1 & 64.1 \\ 
        \bottomrule[1pt]
    \end{tabular}
    }
    \caption{The ablation of attention distillation strategy. '$A_{1,t-v}$' is our first layer $A_{1,t-v}$ only KD, and '$A_{1,v-v}$', '$A_{1,t-t}$' are in the same way. '$A_{1,all}$' refers to applying attention distillation on full attention of the first layer. '$+A_{last,all}$' refers to adding an extra KD loss on full attention of the last layer.}
    \label{table:ablation-attention}
\end{table}

\subsection{Discussions and Limitations}

Align-KD strategy brings benefit to MobileVLM V2 model under both long and short prompt limitations. Align-KD is relatively light designed, which enable possible expansion to resource-limited scenarios. We provide the working expense comparisons in Table \ref{table:resource}, including the total training time and maximum memory occupied during training. The experiments are conducted on 8 Nvidia V100 GPUs. It costs around 296 GPU hours and 676 GPU hours to train Align-KD MVLM2 1.7B on \textit{Short} and \textit{Long} subdatasets. For some examples that would cause instantaneous overload due to the gradient accumulation, we save the output representations from teacher and let them join training later when we can remove the teacher model from the device. When dealing with \textit{Short}, Align-KD can achieve training with maximum memory workload of 22.3 GB per GPU, which is acceptable for commercial devices like RTX 3090. As for \textit{Long} subset, the workload of 30.7 GB per GPU is also acceptable for devices like NVIDIA V100. The memory load of vanilla MobileVLM V2 1.7B is not presented here since it can be flexibly tuned. Despite the extra workload brought, Align-KD still enables universal improvements across different vision-language tasks.
\begin{table}[t]
    \centering
    \setlength{\tabcolsep}{2pt}
    \resizebox{\columnwidth}{!}{
    \begin{tabular}{ccc}
        \toprule[1pt]
        Settings & Total Training Time & Max Memory Occupied\\
        \midrule[0.5pt]
        \textit{Short} MVLM2 1.7B & 176 GPU hours & - \\
        \textit{Long} MVLM2 1.7B & 228 GPU hours & - \\
        \textit{Short} w/ Align-KD & 296 GPU hours & 22.3 GB/device*\\
        \textit{Long} w/ Align-KD & 676 GPU hours & 30.7 GB/device$\dagger$\\
        \bottomrule[1pt]
    \end{tabular}
    }
    \caption{The comparison of training time and memory workload. 'w/ Align-KD' refers to training with Align-KD policy. *Set batch size to 4 per iteration. $\dagger$Set batch size to 1 per iteration.}
    \label{table:resource}
\end{table}

\section{Conclusion}
\label{sec:conclusion}

We propose a knowledge distillation method for MobileVLM V2 model, namely Align-KD in this paper. Based on the conjecture that the alignment mainly happens at the front layer of LLM in VLMs, Align-KD proposes to conduct knowledge distillation only on the text-query-vision part of the first attention. The vision tokens are also unbalancedly enhanced according to the text tokens' focusing. Using MobileVLM V2 7B model as teacher, Align-KD enables universal improvements across benchmarks under both regular training setting and resource-limited setting.  

\section*{Acknowledgment}
\label{sec:ack}

This work was supported by Sichuan Science and Technology Program (No.2024YFHZ0026).
\newpage
{
    \small
    \bibliographystyle{ieeenat_fullname}
    \bibliography{main}
}


\end{document}